\documentclass[sigconf]{acmart}

\usepackage{algorithm}
\usepackage{algpseudocode}
\usepackage{comment}
\usepackage{url}

\setcopyright{none}
\author{Matthew Chak}
\orcid{0000-0002-5659-8932}
\affiliation{%
 \institution{Department of Computer Science}
 \institution{California Polytechnic State University}
 \city{San Luis Obispo, CA}
 \country{US}
}
\email{mchak@calpoly.edu}
\author{Paul Anderson}
\affiliation{%
 \institution{Department of Computer Science}
 \institution{California Polytechnic State University}
 \city{San Luis Obispo, CA}
 \country{US}
}
\email{pander14@calpoly.edu}
\title{Finding Multiple Interpretations in Datasets}
\begin{document}
\begin{abstract}
In this paper, we propose an approach to finding sets of similar-performing models (in terms of loss/accuracy measurements) with highly different context-aware characteristics. Through experiments on the METABRIC dataset, we show that the proposed method finds multiple models with highly different gene expressions than those found by the control methodology without performance penalties. We argue that the proposed methodology is important whenever one aims to analyze any global characteristic of a model to extract insight into the underlying phenomenon being studied.
\end{abstract}

\maketitle

\section{Introduction}
In many contexts, especially biology, characteristics of deep-learning models are often examined to derive insight about the underlying phenomenon being studied \cite{dlXAiNPG}. This can take many forms, i.e. using feature importance metrics to discover important predictors of a phenomenon of interest (\cite{cancers15133463}, \cite{plosGlobalImportance}, \cite{naturePermutationbasedIdentification}), using neural net representation layers to discover patient clusters (\cite{nihDeeplearningApproach}, \cite{natureDeepPatient}), and more. In general, these studies loosely follow a three-step approach:

\begin{enumerate}
    \item Train a deep-learning model that achieves satisfactory performance on both both training and held-out testing data
    \item Extract some global property of the model (e.g. feature importance, representation layers, saliency maps for CNNs, etc.)
    \item Examine said property to extract a general context-specific conclusion
\end{enumerate}

Though the above approach is philosophically sound \cite{arxivDefencePosthoc}, conclusions derived in the above manner are valid only if no other alternative explanations with similar explanatory power exist. For example, if a biologist uses a model's feature importance to conclude that gene X is critical in predicting cancer, the conclusion is only valid if there don't exist other models with similar performance where gene X's feature importance is low (\cite{Harman1965}, \cite{Veit2019}). 

Due to the high-dimensional nature of many biological datasets, however, underspecification is often a possibility \cite{DAmour2020}, meaning such a uniqueness-of-explanation condition may not hold. It is thus incumbent upon researchers to ensure that they have fully explored their model space before asserting a domain-specific conclusion. This paper introduces a method to allow researchers to do such an exploration.

\section{Methodology}
From a technical perspective, our work addresses the following problem: Given a set of models $\{\theta_1, ..., \theta_n\}$, can we produce a new model $\theta_{n + 1}$ that is in some ways "different" from all of the models in the set and yet has similar metrics when evaluated on the training and testing data (i.e. test and train accuracies)? Of course, what it means to be "different" is context-specific, but for our purposes, we view the difference in the context of some function $\delta: \Theta \times \Theta \to \mathbb{R^+}$. Our task then becomes solving the following optimization problem:

\begin{equation}
\begin{cases}
    \textbf{maximize } \sum_{i=1}^{n}\delta(\theta_{i}, \theta_{n+1}) & \\
    \textbf{subject to } \lambda(\theta_{n+1}) < \lambda(\theta_{i}) + \varepsilon
\end{cases}
\end{equation}

Where $\lambda$ is our primary loss and $\varepsilon$ is some acceptable tolerance. When $\delta$ is non-differentiable, such a problem can be solved using proxy-constrained optimization approaches (\cite{Cotter2018}, \cite{gallegoPosada2022cooper}). For this paper, however, we deal with differentiable distance functions only, meaning our formulation can be reduced to a form much more easily optimizable in a deep-learning setting:

\begin{equation}
    \textbf{minimize} \quad \lambda(\theta_{n + 1}) + \alpha\sum_{i=1}^{n}\frac{1}{1 + \delta(\theta_{i}, \theta_{n+1})}
\end{equation}

Suppose now that we have a set of $n$ existing models with respective parameters \begin{math}\{\theta_0, \theta_1, ..., \theta_{n - 1}\}\end{math}. To obtain a new model with (hopefully) both a decent performance and a different interpretation from all the other models in a set, we use the following algorithm:

\begin{algorithm}[H]
\caption{NextModel}
\label{alg:conf}
\begin{algorithmic}
\Require A set of models with parameters \begin{math}\{\theta_0, \theta_1, ..., \theta_{n - 1}\}\end{math}, a differentiable loss function $\lambda: \Theta \to \mathbb{R}$, a numbers of epochs $T$, a learning rate $\eta$, and a weight $\alpha$
\Ensure A new model with parameters $\theta_{n}$
\State Initialize $\theta_{n}$ randomly
\For {$t \in [T]$}:
\State Let $J_{perm}^{(t)}$ be $\frac{1}{n}\sum_{i=0}^{n - 1}\frac{1}{1 + \delta(\theta_{i}, \theta_{n})}$
\State Let $\Delta^{(t)}$ be the gradient of $(\lambda(\theta_{n}) + \alpha J_{perm}^{(t)})$ w.r.t. $\theta_{n}$
\State Update $\theta_{n}^{(t + 1)} = \theta_{n}^{(t)} - \eta\Delta^{(t)}$
\EndFor
\State \Return $\theta_n^T$
\end{algorithmic}
\end{algorithm}

More simply, we train a model in the typical fashion using $\frac{1}{1 + \delta(\theta_i, \theta_n)}$ as a weighted regularization term, thereby penalizing models with similar interpretations. This procedure is then repeated until a suitable number of models is generated. For performance reasons, one can choose to limit the number of models considered in the set of distant models, i.e. by utilizing a sliding window.

\section{Experiment}
\subsection{Methodology}
We first trained 10 DeepType models on the METABRIC dataset \cite{Curtis2012} using the procedure outlined in \cite{nihDeeplearningApproach}, whose authors kindly converted genetic probe data into gene expression data necessary for our analysis. Training was done using the `torch-deeptype` package \cite{PhysBoom2025}. The hyperparameters in \cite{nihDeeplearningApproach} were modified slightly to $\lambda = 0.003, \alpha = 0.2$ instead of $\lambda = 0.006, \alpha = 1.2$ as we found that the original hyperparameters led to only a small percentage of control models achieving satisfactory performance (whereas all models converged to an adequate solution with the new hyperparameters). The best model checkpoints as measured by the combined loss described in \cite{nihDeeplearningApproach} on validation data were kept. The importances of each gene (as measured by input layer weights) were then obtained and ranked for each model. 

For the distance function, the most important genes for the new model were found using a differentiable ranking algorithm (\cite{blondel2020fast}) on the input layer weights, and the presence of any gene in the top 25 of existing models in the top 25 of the new model was penalized. Distance was thus computed as follows:

\begin{algorithm}[H]
\caption{RankLossDistance}
\label{alg:conf}
\begin{algorithmic}
\Require A reference model $\theta_{old}$, A new model $\theta_{new}$, and a number of genes $n$ (for us, 25)
\Ensure The distance $\delta(\theta_{old}, \theta_{new})$
\State Compute 
  $orig = \text{softrank}\bigl(\texttt{input\_weights}(\theta_{\text{old}})\bigr)$
\State Find input indices of the top $n$ important inputs as $indices$
\State Compute $offset = -orig[indices[n]]$, the softrank importance of the $n + 1$'th most important gene
\State Compute $new = \text{softrank}\bigl(\texttt{input\_weights}(\theta_{\text{new}})\bigr)$
\State \Return $\sum_{i = 0}^{n - 1} max((new[indices[i]] - offset) \cdot (n - i), 0)$
\end{algorithmic}
\end{algorithm}

Effectively, we first compute the softrank of the input importances of the original model. We then find the most important genes by performing an argsort on that softrank. We then take the importance of the 26'th gene and store it as the offset. This way, $importance - offset > 0$ for any gene ranked above the 26'th, and is less than 0 for any below it. We then compute the importances for the new model and softrank them. Though not entirely accurate, any gene with a softrank above the previously computed offset should be in roughly the top 25 important genes for the new model, with a higher positive difference between $new - offset$ indicating a higher rank, and a negative difference indicates the gene is roughly outside of the top 25 genes. As such, the distance is computed as a weighted sum of the positive differences between the ranks of the top 25 genes of the old model in the new model and the offset. This distance function is completely differentiable and an implementation can be found in the code for this experiment \cite{Chak2025}.

Using (Algorithm 1) along with this distance functions, 3 new models were trained using a distance weight of 0.01. To start, one well-performing control model from an earlier run using \cite{nihDeeplearningApproach}'s original hyperparameters was used - not using one of the 10 control models produced earlier was an oversight. Each time a new model was trained, it was added to the list of models passed into (Algorithm 1) for the next model. Important gene expressions and test/train/validation metrics as well as patient clusters were recorded for all models

\subsection{Results}
Summary statistics for the 10 control models can be found in this table:

\begin{table}[ht]
  \centering
  \caption{Summary statistics for the 10 control models}
  \label{tab:control-model-stats}
  \begin{tabular}{lrrrr}
    \toprule
    Metric & Mean & Std & Min & Max \\
    \midrule
    Train Accuracy & 0.946971 & 0.029941 & 0.890554 & 0.988925 \\
    Validation Accuracy   & 0.804678 & 0.013561 & 0.777778 & 0.830409 \\
    Test Accuracy  & 0.803981 & 0.020418 & 0.775176 & 0.836066 \\
    Primary Loss   & 0.294124 & 0.050043 & 0.220923  & 0.388864  \\
    Weighted Sparsity Loss  & 0.288114 & 0.007658 & 0.275222  & 0.299920  \\
    Weighted Cluster Loss   & 0.085172 & 0.029122 & 0.035912  & 0.127430  \\
    \bottomrule
  \end{tabular}
\end{table}

For the three models trained with algorithm 1 (listed in sequential order of creation), the metrics are below:

\begin{table}[ht]
  \centering
  \caption{Comparison of best accuracies and loss components for the three models}
  \label{tab:model-comparison}
  \begin{tabular}{lrrrrrr}
    \toprule
    Model   & Train Acc & Val Acc & Test Acc & Primary    & Sparsity   & Cluster    \\
    \midrule
    Model 1 & 0.9537         & 0.8187       & 0.8220        & 0.2841     & 0.2973     & 0.0809     \\
    Model 2 & 0.9922         & 0.8304       & 0.7822        & 0.2329     & 0.3142     & 0.0440     \\
    Model 3 & 0.9993         & 0.7719       & 0.8056        & 0.1752     & 2.0578     & 0.1521     \\
    \bottomrule
  \end{tabular}
\end{table}

As seen, all models exhibited reasonable performance on all metrics relative to the control models, except model 3 which arguably exhibited an unacceptable sparsity loss. 

When examining the gene expressions of the control models, across all ten models, 65 genes appeared in the top 25 genes by input weight, with nine genes appearing in the top 25 for all ten models. For models 1, 2, and 3, however, the top 25 genes of each model were unique - that is, 75 genes were found across the top 25 genes for each of the three models, which is the maximum possible. Of these 75, only 26 were found in the top 25 gene expressions of any control model. Full gene expression data can be found in the supplementary data.

\section{Discussion}
Our results confirm that, even on a well-studied dataset like METABRIC, there exist clusters of models that achieve equivalent predictive performance while relying on entirely different sets of genes.  In particular, the control models shared a small set of common top genes across all ten runs, whereas each of the three “distant” models introduced a wholly unique top-25 gene list for each model.  This underscores two points: first, relying on a single “best” model for biological interpretation risks missing alternative but equally plausible mechanisms; second, our regularized training with a differentiable distance term is an effective way to surface these alternatives without compromising overall accuracy.

As a word of caution, the proposed method will generally lead to worse-performing models over time, especially in datasets where there is truly one "optimal" solution. In our results, this is seen with model 3, where optimizing distance came at the cost of sparsity. This, however, is not necessarily a negative, as diminishing metrics simply indicate that the majority of feasible solutions have been discovered. As such, we recommend repeating the process described until metrics start decreasing noticeably.

\section{Conclusion}
We have presented a scalable method for discovering multiple, high-performing neural models with maximally different interpretations. On the METABRIC dataset, we generated three alternative models whose gene-importance profiles were entirely disjoint from those of the ten baseline runs, illustrating both the feasibility and necessity of our approach. Given the generality and ease of use of our method, we believe it should be an integral part of any workflow aiming to derive domain insight from global characteristics of models.

\section{Future Work}
Directions for future work are plentiful:

\begin{enumerate}
    \item The proposed methodology could be used on a wide array of datasets throughout biology and elsewhere to extract interesting insights.
    \item A result that is both intuitively plausible and that is somewhat supported in our experiments is that, as more models get added to the input into (Algorithm 1), primary metrics will tend to decrease. The rate at which they decrease may be an interesting measure of the level of underspecification of a machine learning pipeline: if no alternative solution is found without severe penalties, then underspecification may not be present, whereas if metrics remain high for many iterations of the algorithm, the pipeline may be underspecified. This link should be investigated.
    \item So far, experimentation has been entirely emperical. Are there theoretical bounds that could be derived for the process described, i.e. in estimating the error-distance tradeoff or in estimating the number of models greater than some threshold away from all models in the set that are still within reasonable error?
    \item Could the $\alpha$ described in (Algorithm 1) be dynamically selected, e.g. via constrained optimization?
\end{enumerate}

The list is not exhaustive, but is hopefully a good start.

\bibliographystyle{ACM-Reference-Format}
\bibliography{bibliography.bib}

@Article{Cotter2018,
  author     = {Andrew Cotter and Heinrich Jiang and Karthik Sridharan},
  journal    = {CoRR},
  title      = {Two-Player Games for Efficient Non-Convex Constrained Optimization},
  year       = {2018},
  volume     = {abs/1804.06500},
  bibsource  = {dblp computer science bibliography, https://dblp.org},
  comment    = {Cotter's PC},
  eprint     = {1804.06500},
  eprinttype = {arXiv},
  ranking    = {rank5},
  url        = {http://arxiv.org/abs/1804.06500},
}

@Article{DAmour2020,
  author     = {Alexander D'Amour and Katherine A. Heller and Dan Moldovan and Ben Adlam and Babak Alipanahi and Alex Beutel and Christina Chen and Jonathan Deaton and Jacob Eisenstein and Matthew D. Hoffman and Farhad Hormozdiari and Neil Houlsby and Shaobo Hou and Ghassen Jerfel and Alan Karthikesalingam and Mario Lucic and Yi{-}An Ma and Cory Y. McLean and Diana Mincu and Akinori Mitani and Andrea Montanari and Zachary Nado and Vivek Natarajan and Christopher Nielson and Thomas F. Osborne and Rajiv Raman and Kim Ramasamy and Rory Sayres and Jessica Schrouff and Martin Seneviratne and Shannon Sequeira and Harini Suresh and Victor Veitch and Max Vladymyrov and Xuezhi Wang and Kellie Webster and Steve Yadlowsky and Taedong Yun and Xiaohua Zhai and D. Sculley},
  journal    = {CoRR},
  title      = {Underspecification Presents Challenges for Credibility in Modern Machine Learning},
  year       = {2020},
  volume     = {abs/2011.03395},
  bibsource  = {dblp computer science bibliography, https://dblp.org},
  eprint     = {2011.03395},
  eprinttype = {arXiv},
  url        = {https://arxiv.org/abs/2011.03395},
}

@misc{gallegoPosada2022cooper,
    author={Gallego-Posada, Jose and Ramirez, Juan},
    title={{Cooper: a toolkit for Lagrangian-based constrained optimization}},
    howpublished={\url{https://github.com/cooper-org/cooper}},
    year={2022}
}

@misc{nihDeeplearningApproach,
	author = {Chen, Runpu and Yang, Le and Goodison, Steve and Sun, Yijun},
	title = {Deep-learning approach to identifying cancer subtypes using high-dimensional genomic data},
	url = {https://pubmed.ncbi.nlm.nih.gov/31603461/},
	year = {2020},
}

@misc{plosGlobalImportance,
	author = {Koo, Peter and Majdandzic, Antonio and Ploenzke, Matthew and Anand, Praveen and Paul, Steffan},
	title = {Global importance analysis: An interpretability method to quantify importance of genomic features in deep neural networks},
	url ={https://journals.plos.org/ploscompbiol/article?id=10.1371%2Fjournal.pcbi.1008925},
	year = {2021},
}

@misc{dlXAiNPG,
	author = {Wang, Chenyu and Zuo, Chaoying and Su, Zihan and Xing, Yuhang and Li, Lu and Wang, Maojun and Zhang, Zeyu},
	title = {Deep Learning and Explainable AI: New Pathways to Genetic Insights},
	url = {https://arxiv.org/html/2505.09873v1},
	year = {2025},
}

@misc{naturePermutationbasedIdentification,
	author = {Mi, Xinlei and Zou, Baiming and Zou, Fei and Hu, Jianhua},
	title = {Permutation-based identification of important biomarkers for complex diseases via machine learning models},
	url = {https://www.nature.com/articles/s41467-021-22756-2},
	year = {2021},
}

@misc{cancers15133463,
	author = {Cheon, Wonjoong and Han, Mira and Jeong, Seonghoon and Oh, Eun Sang and Lee, Sung Uk and Lee, Se Byeong and Shin, Dongho and Lim, Young Kyung and Jeong, Jong Hwi and Kim, Haksoo and Kim, Joo Young},
	title = {Feature Importance Analysis of a Deep Learning Model for Predicting Late Bladder Toxicity Occurrence in Uterine Cervical Cancer Patients},
	url={https://www.mdpi.com/2072-6694/15/13/3463},
	year = {2023},
}

@misc{natureDeepPatient,
	author = {Miotto, Riccardo and Li, Li and Kidd, Brian and Dudley, Joel},
	title = {Deep Patient: An Unsupervised Representation to Predict the Future of Patients from the Electronic Health Records},
	url = {https://www.nature.com/articles/srep26094},
	year = {2016}
}

@misc{arxivDefencePosthoc,
	author = {Oh, Nick},
	title = {In Defence of Post-hoc Explainability},
	url = {https://arxiv.org/abs/2412.17883},
	year = {2024},
}

@article{Harman1965,
  author       = {Harman, Gilbert},
  title        = {The Inference to the Best Explanation},
  journal      = {The Philosophical Review},
  volume       = {74},
  number       = {1},
  pages        = {88--95},
  year         = {1965},
  doi          = {10.2307/2182135},
}

@misc{Veit2019,
  author       = {Veit, Walter},
  title        = {Model Pluralism},
  howpublished = {arXiv preprint arXiv:1909.13653},
  year         = {2019},
  url          = {https://arxiv.org/abs/1909.13653},
}

@misc{PhysBoom2025,
  author       = {Matthew Chak},
  title        = {torch-deeptype},
  url = {https://github.com/PhysBoom/torch-deeptype},
  year         = {2025},
}

@article{Curtis2012,
  author       = {Curtis, C. and Shah, S.-P. and Chin, S.-F. and Turashvili, G. and Rueda, O. M. and Dunning, M. J. and Speed, D. and Lynch, A. G. and Samarajiwa, S. and Yuan, Y. and Graf, S. and Ha, G. and Haffari, G. and Bashashati, A. and Russell, R. and McKinney, S. and METABRIC Group and others},
  title        = {The Genomic and Transcriptomic Architecture of 2,000 Breast Tumours Reveals Novel Subgroups},
  journal      = {Nature},
  volume       = {486},
  number       = {7403},
  pages        = {346--352},
  year         = {2012},
  doi          = {10.1038/nature10983},
}

@inproceedings{blondel2020fast,
  title={Fast differentiable sorting and ranking},
  author={Blondel, Mathieu and Teboul, Olivier and Berthet, Quentin and Djolonga, Josip},
  booktitle={International Conference on Machine Learning},
  pages={950--959},
  year={2020},
  organization={PMLR}
}

@misc{Chak2025,
  author       = {Chak, Matthew},
  title        = {deeptype-push-apart},
  url ={https://www.kaggle.com/code/matthewchak/deeptype-push-apart},
  year         = {2025},
  note         = {Kaggle notebook; accessed 2025-06-11},
}

\end{document}